\documentclass[letterpaper, 10 pt, conference]{ieeeconf}  

\IEEEoverridecommandlockouts                              

\usepackage{amsmath}    
\usepackage{amssymb}
\usepackage{graphicx}   
\usepackage{verbatim}   
\usepackage{color}      
\usepackage{subfigure}  
\usepackage{hyperref}   

\newcommand{\beq}{\begin{equation}}
\newcommand{\eeq}{\end{equation}}
\newcommand{\beqa}{\begin{eqnarray}}
\newcommand{\eeqa}{\end{eqnarray}}
\newcommand{\mv}{\mathbf{m}}
\newcommand{\iv}{\mathbf{u}}
\newcommand{\tv}{\mathbf{t}}
\newcommand{\ms}{\mathcal{M}}
\newcommand{\Nl}{L}
\newcommand{\Nm}{N_m}
\newcommand{\coord}{\mathbf{c}}

\title{\LARGE \bf
Landmark2Vec: An Unsupervised Neural Network-Based Landmark Positioning Method
}

\author{Alireza Razavi$^{1}$
\thanks{$^{1}$Alireza Razavi is with Autonomous Transport Solutions (ATS), Scania CV AB, S\"{o}dert\"{a}lje, Sweden,
        {\tt\small s.a.razavi@gmail.com, alireza.razavi@scania.com}. The work was partly funded by Sweden’s innovation agency, Vinnova, under project number 2018-02700.}}

\begin{document}

\maketitle
\thispagestyle{empty}
\pagestyle{empty}

\begin{abstract}

A Neural Network-based method for unsupervised landmarks map estimation from  measurements taken from landmarks is introduced. The measurements needed for training the network are the signals observed/received from landmarks by an agent. The definition of landmarks, agent, and the measurements taken by agent from landmarks is rather broad here: landmarks can be visual objects, e.g., poles along a road, with measurements being the size of landmark in a visual sensor mounted on a vehicle (agent), or they can be radio transmitters, e.g., WiFi access points inside a building, with measurements being the Received Signal Strength (RSS) heard from them by a mobile device carried by a person (agent). The goal of the map estimation is then to find the positions of landmarks up to a scale, rotation, and shift (i.e., the topological map of the landmarks). Assuming that there are $\Nl$ landmarks, the measurements will be $\Nl \times 1$ vectors collected over the area. A shallow network then will be trained to learn the map without any ground truth information.

\end{abstract}


\section{Introduction} 
Localization is an important task for autonomous vehicles and robots as well as for hand-held mobile devices with Location-Based Services (LBS). Finding the agent pose using GNSS-based localization methods is not always possible for various reasons: in indoor and closed areas where LBS are of high interest, the GNSS signals cannot be received. Besides, in dense urban scenarios, multipath effect usually deteriorates the accuracy of GNSS receivers, and even when there is reception, the accuracy of consumer GNSS-based devices is not good enough for sensitive and safety-critical tasks such as autonomous driving in which we need sub-meter accuracy. The more accurate devices like dGPS or RTK-GPS \cite{gps1,gps2} are too expensive to be equipped on every passenger car or mobile device. This motivates the need for landmark-based localization in which we use the coordinates of landmarks plus distances of agent to landmarks to estimate the agent's pose. To this end, however, we need to have a map of the area including the (accurate) coordinates of the landmarks. 

In practice, however, the landmarks locations are not known beforehand and needed to be estimated themselves first. This is usually done based on the signals received/observed from them by an agent. Conventional data-driven methods for estimating the coordinates of landmarks need to collect measurements from landmarks together with ground-truth information (being the coordinates of measurement locations) in order to estimate the landmarks locations. For instance, in case of WiFi localization this can be done by methods such as Weighted Centroid Localization (WCL) \cite{wcl1,wcl2} or using parametric methods based on propagation model \cite{wifi1, wifi2}. Collecting ground-truth labeled data is very often time consuming, expensive, and labor-intensive. For instance, to collect ground-truth coordinates of measurement points in the outdoor environments, we need expensive GNSS receivers with high accuracy. 
In indoor localization, e.g., WiFi-based localization, in order to estimate the location of access points, the data collection is carried out labor intensively by recording the coordinates at which the measurements are collected in addition to the RSS heard by WiFi receiver carried by the agent at each location.

To overcome this problem, in this paper a method is proposed for unsupervised landmark localization up to a rotation, transformation, and scale, or in other words, the topological map of landmarks. We design and train a network which learns the relative positions of all landmarks. The idea is akin to the celebrated Word2Vec idea for word embedding \cite{word2vec1}, and therefore we call it Landmark2Vec.

{\bf Paper organization: } The rest of this paper is organized as follows: In Section \ref{formulation} the problem and its assumptions are explained and formulated. Section \ref{landmark2vec} is the core part of the paper where the Landmark2Vec method is introduced in details. In Section \ref{evaluation}, we provide a metric for performance evaluation of the proposed method in terms of map reconstruction and propose a heuristic stopping criterion for training. Some numerical experiments is provided in Section \ref{numeric}. Finally the paper is concluded in Section \ref{conclusion}.

{\bf Notation: } For the sake of consistency, we adopt the following notations throughout this paper: $\triangleq$ denotes definition while $=$ denotes equation. Vectors are denoted by small boldface letters (e.g., $\mathbf{v}$) and matrices are denoted by capital boldface letters (e.g., $\mathbf{M}$). $\| . \|$ denotes the second norm. Coordinates are shown by boldface small letter $\coord\triangleq(x,y)$. To distinguish between the coordinates of landmarks and agent, we denote the coordinates of agent at location $i$ by using subscript $\coord_i\triangleq(x_i,y_i)$ and the coordinates of landmark $l$ by using superscript $\coord^l\triangleq(x^l, y^l)$. 

\section{Problem Formulation}
\label{formulation}
Consider an area deployed with $\Nl$ landmarks and the ultimate goal is to use the measurements/signals from landmarks for positioning. But, to this end, we first need to know the positions of landmarks. Without loss of generality, we assume that all the coordinates are two-dimensional (2-D). In conventional methods, to find the position of landmarks, we collect a relatively large number of measurements, say $\Nm$, from landmarks in the area. Let denote the set of measurements by:
\beq
\ms \triangleq \{\coord_i, \mv_i\}_{i=1}^{\Nm},\label{meas_set}
\eeq
where $\coord_i \triangleq(x_i,y_i)$ denotes the 2-D coordinates of $i$-th measurement, and $\mv_i \triangleq [m_{i,1},m_{i,2},\ldots,m_{i,\Nl}]$ is the measurement vector taken at $\coord_i$, whose $l$-th entry $m_{i,l}$ is the observation from landmark $l$. This can be any signal which is (not necessarily linearly) proportional to the distance of the landmark to the pose $(x_i,y_i)$; it can be the range from a pole (landmark) along the road measured by a lidar sensor or camera, or the received signal strength (RSS) heard from a WiFi transmitter (landmark) by a mobile device.

Conventional methods use the ground-truth location information, i.e., coordinates $\{\coord_i \triangleq (x_i,y_i)\}_{i=1}^{\Nm}$, together with the measured signals $\{\mv_i\}_{i=1}^{\Nm}$, to estimate the positions of landmarks. In the following we review two existing approaches for landmarks and agent pose estimation. 

\subsection{Weighted Centroid Localization}
\label{seq:wcl}
The weighted centroid localization (WCL) approach, first proposed for position estimation in
wireless sensor networks \cite{wcl1, razavi2015k}, is a simple and low-complexity yet promising localization approach. One of the main advantages of WCL is that it is not needed to first convert the measurements to distance. This is especially useful when the relationship between the measurements and distance is complicated and noisy, e.g., through pathloss model (see equation (\ref{eq:pathloss})). 

The WCL method \cite{wcl1,wcl2,razavi2015k} estimates the coordinates of a landmark, say $l$-th landmark, using the following formula:
\beq
\hat{\coord}^l = \frac{\sum_{i=1}^{\Nm} m_{i,l}\coord_i}{\sum_{i=1}^{\Nm} m_{i,l}}.
\label{eq:wcl1}
\eeq
Then, having all the estimates $\{\hat{\coord}^l\}_{l=1}^{\Nl}$, during the test (agent positioning) phase, if an agent receives a measurement vector $\mv_j \triangleq [m_{j,1},m_{j,2},\ldots,m_{j,\Nl}]$ at a pose $j$, the coordinates of $j$ can be estimated as \footnote{We remark again that estimation of agent's pose is not the subject of the current paper. However, we explain it here to better motivate the need for finding landmarks positions.}
\beq
\hat{\coord}_j = \frac{\sum_{l=1}^{\Nl} m_{j,l}\hat{\coord}^l}{\sum_{j=1}^{\Nl} m_{j,l}}.
\label{eq:wcl2}
\eeq

\subsection{Multilateration}
\label{sec:lateration}
If it is easy to convert the sensor measurements to distances, then multilateration is a promising technique for positioning \cite{lateration1, lateration2, lateration3, lateration4}. This is the case for example when we use a Lidar or a camera for range measurement. Once the distance of a given landmark is known to three or more points with known coordinates (i.e., some ground-truth information) then we can use the multilateration technique to estimate the location of that landmark; see, e.g., \cite{lateration2}.

We re-emphasize here that in both WCL and Multilateration, as well as other conventional techniques, the ground-truth information is required for performing landmark positioning.

Next, we introduce our proposed unsupervised method for landmark positioning which estimates the topological map of landmarks only from measurements $\{\mv_i\}_{i=1}^{\Nm}$.

\section{Landmark2Vec Method}
\label{landmark2vec}
In this section, the Landmark2Vec method is presented. The general NN architecture is introduced, the data collection and preparation for training is explained, the training is presented, and then we will describe how the trained model can be used for landmark positioning. The connection of the method to the problems of word embedding and graph embedding will be also described briefly. 

\subsection{Architecture}
\label{architecture}
\begin{figure}[t!]
	\centering
		\includegraphics[scale=0.65]{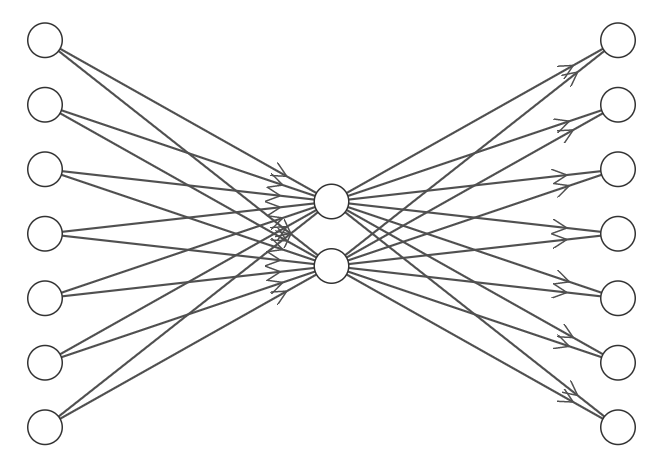}
	\caption{Landmark2Vec network architecture: The number of neurons in the input and output layers both equal the number of landmarks. The number of neurons in the middle layer is 2 for 2-dimensional localization and 3 for 3-dimensional localization. The activation functions of the middle layer is linear and of output layer is softmax.}
	\label{fig:nn}
\end{figure}

The general architecture for the network is as follows:
\begin{enumerate}
\item The neural network architecture used for unsupervised landmark localization is a butterfly shaped fully-connected network as depicted in Fig \ref{fig:nn}.

\item The number of input neurons equals the number of landmarks.
\item The number of output neuron also equals the number of landmarks.
\item The number of neurons in the middle layer (bottleneck) equals the number of dimensions (2 if we are doing 2-dimensional localization, 3 if we are doing 3-dimensional localization).
\item The activation layer of the middle layer is linear.
\item The activation functions of the output layer should be softmax so as to generate numbers between 0 and 1 which sum up to 1.

\end{enumerate}

\subsection{Data Collection, Preparation, and Training}
\label{sec:collection}

To collect data for training the network, we measure the signals/observations from all landmarks at as many locations as possible in the area of interest. Assuming that there are $\Nl$ landmarks, at each location we measure $\Nl$ signals/observations, one from each landmark, which can be put in an $\Nl$-dimensional vector, where the first element of the vector is the signal received or observed from landmark number $1$, the second element of the vector is the signal received or observed from landmark number $2$, and so on and so forth.
I emphasize again that we do not need to record the coordinates of the measurement points as the proposed method is an unsupervised one. 
This will hugely reduce the cost and labor time needed for landmark positioning, which is the main advantage of the proposed method.

As mentioned in subsection \ref{architecture}, the number of neurons in both input layer and output layer equal the number of landmarks, i.e., $\Nl$. Therefore, to train the network, we must provide $\Nl$-dimensional vectors to feed both as input to the input layer and as target to the output layer.
We build both input and target vectors from the collected data vectors. In other words, from each $\Nl$-dimensional collected data vector we build one $\Nl$-dimensional input vector and one $\Nl$-dimensional target vector. Denoting a measurement by $\mv_i \triangleq [m_{i,1},m_{i,2},\ldots,m_{i,\Nl}]$, in order to build an input vector $\iv_i$ and a target vector $\tv_i$ from it, we follow the following steps:
\begin{enumerate}
	\item We choose a number $n$, where $2\le n \le \Nl$. This is akin to $n$ in $n$-gram\footnote{An $n$-gram is a sequence of $n$ words in a sentence of text or speech.} concept in Natural Language Processing (NLP), and basically is used to only select $n$ landmarks whose measured values are more similar at each measurement, or in other words are more correlated location-wise. From $\mv_i$, we then keep $n$ largest (in magnitude) entries and set the rest equal to zero. Let call this vector $\tilde{\mv}_i$. 
	\item Input vector $\iv_i$ is an $\Nl$-dimensional vectors with one element equals $1$ and the other $\Nl-1$ elements equal $0$. The only $1$ element is where the collected data vector $\mv_i$ has its largest value.
	\item Now to build the target vector $\tv_i$, we set the largest entry of $\tilde{\mv}_i$ to zero (i.e., the one corresponding to the only nonzero element of input vector $\iv_i$) and then normalize the other $n-1$ nonzero entries of $\tilde{\mv}_i$ such that they sum up to 1. 
\end{enumerate}

To provide a toy example, assume that there are $\Nl=6$ landmarks and at a (unknown) location the $6$-dimensional measured vector is $\mv_i=[1,2,8,4,3,1]$. Assuming that we have chosen $n=4$, then $\tilde{\mv}_i=[0,2,8,4,3,0]$. The input vector built from $\mv_i$  will be $\iv_i=[0,0,1,0,0,0]$ and the $6$-dimensional target vector will be $\tv_i=[0,\frac{2}{9},0,\frac{4}{9},\frac{3}{9},0]$. 

During the training, we fed the input layer by the input vectors $\iv_i$ built as described above. The network then spits out an output vector in its output layer. The loss function then will be the cross entropy between this ``output vector'' and the corresponding ``target vector'' $\tv_i$ built from training data as explained above. The training is done through the back-propagation algorithm.

\subsection{Inference}
\label{inference}

After the training is finished, the last step is to infer the location of landmarks from the trained model.
In order to find the location of a landmark, say landmark $l$, we feed the trained network by a vector whose $l$-th element equals 1 and the other $\Nl-1$ elements are $0$. What network generates in response to this vector in the output of its middle layer (bottleneck), is the coordinates of landmark $l$. If we are doing the localization in $2$-dimensional space, then there will be $2$ neurons in the middle layer. The first one represents $x$-coordinate of the landmark's location and the second one represents the $y$-coordinate. Similarly, if the space is $3$-dimensional, there will be three neurons in the middle layer whose output will be $(x,y,z)$ coordinates of the landmarks.

Equivalently, we can say that the trained weight between $l$-th input neuron and the first middle neuron is the $x$ coordinate of landmark $l$ and the weight between it and the second middle neuron is the $y$-coordinate of landmark $l$.

\subsection{Analogies between Word2Vec and Landmark2Vec}
\label{analogies}
To shed some more light on the proposed landmark2vec method, in this section we provide some analogies between Landmark2Vec and the celebrated Word2Vec algorithm for word embedding in NLP \cite{word2vec1}. Word2Vec exploits the co-occurrences of words in $n$-grams to find their relative positions in an embedded space. Similarly in Landmark2Vec, the relative position of landmarks is inferred based on the similarity of their values in recorded measurements.  

The ingredients of the two algorithms have analogies as described below \footnote{There are two main variants for word2vec: Continuous Bag Of Words (CBOW) and Skip-gram. The one analogous to the proposed landmark2vec algorithm here is Skip-gram.}:

\begin{itemize}
	\item A ``landmark'' in Landmark2Vec is analogous to a ``word'' in Word2Vec. Both are to be embedded to a low-dimensional space.
	\item The number of distinct words in corpus in Word2Vec is analogous to the number of landmarks $\Nl$ in Landmark2Vec.
	\item A measurement is analogous to a sentence.
	\item The $n$-gram role is played by the $n$ nonzero elements of vector $\tilde{\mv}_i$ as defined in subsection \ref{sec:collection}. The diffrence is that in Landmark2Vec we only have one $n$-gram per measurement while in Word2Vec we can have multiple $n$-grams per sentence.
	\item The center word in $n$-gram is like the strongest landmark in terms of received/observed signal in a measurement.
	\item Context words are analogous to the other $n-1$ nonzero landmarks in $\tilde{\mv}_i$.
\end{itemize}

\subsection{Connection to Graph Neural Networks}
\label{GNN}

Graph Neural Networks (GNN) is the extension of Neural Networks for graph data \cite{deepwalk, gnn1, gnn2, node2vec}. Graph embedding is the task in GNNs where the goal is to embed a graph or part of it (a subgraph, subset of nodes, or subset of links) into a low-dimensional space. Landmark2Vec can also be formulated as a graph embedding problem in the following way: 

The nodes of the graph are all the measurements plus the landmarks. In other words, the graph has $\Nm+\Nl$ nodes. Each of the $\Nl$ nodes corresponding to landmarks is connected with a link to all $\Nm$ measurement nodes, where the weight of a link between a landmark node and a measurement node is the received/observed signal form that landmark in that measurement. There is no links between the measurement nodes, and also there is none between the landmark nodes.
The Landmark2Vec can be then thought of as embedding a subset of nodes, namely the nodes corresponding to $\Nl$ landmarks, to a 2-dimensional space.

\section{Evaluation Metric and Stopping Criterion}
\label{evaluation}
In this section, a metric for evaluating the proposed algorithm is introduced and based on that discuss the overfitting problem and suggest a heuristic stopping criterion for avoiding overfitting.

\subsection{Sum of Squared Matching Errors}
\label{sec:ssme}

As mentioned earlier when introducing the proposed method, the output of the proposed algorithm is a topological map of landmarks, i.e., landmarks positions, up to a scale, translation, and rotation. In other words, what the method retrieves from the landmarks measurements is the {\it relative} positions of landmarks.
To compare the true landmarks locations and the estimated scaled-translated-rotated landmarks locations, we need a metric which is independent of scale, location, and translation.

Without loosing the generality, let assume that we are carrying out the localization in $2$-D space. Denoting the true $2$-D coordinate vector of $l$-th landmark by $\begin{bmatrix} x^l,y^l\end{bmatrix}^T \in \mathbb{R}^2$ and the estimated coordinate vector by $\begin{bmatrix} \hat{x}^l,\hat{y}^l\end{bmatrix}^T \in \mathbb{R}^2$, in case of perfect recovery, we expect these two to be related through the following equation:
\begin{eqnarray}
\begin{bmatrix} x^l\\y^l\end{bmatrix}=\mathbf{A} \begin{bmatrix} \hat{x}^l \\ \hat{y}^l\end{bmatrix} + \mathbf{b},
\label{eq:0}
\end{eqnarray}
where $\mathbf{A}\in \mathbb{R}^{2\times 2}$ accounts for the rotation and scale, and $\mathbf{b} \in \mathbb{R}^{2 \times 1}$ accounts for the translation. An imperfect recovery then can be represented by an error term $\mathbf{w}$ which we call the recovery error. Equation (\ref{eq:0}) then can be rewritten as:
\begin{eqnarray}
\begin{bmatrix} x^l\\y^l\end{bmatrix}=\mathbf{A} \begin{bmatrix} \hat{x}^l \\ \hat{y}^l\end{bmatrix} + \mathbf{b} + \mathbf{w},
\end{eqnarray}
 To evaluate the method we then use the following Sum of Squared Matching Errors (SSME) metric:
\beqa
\mathrm{SSME} = \sum_{l=1}^{\Nl} \Big\| \begin{bmatrix} x^l\\y^l\end{bmatrix} - \mathbf{\hat{A}} \begin{bmatrix} \hat{x}^l \\ \hat{y}^l\end{bmatrix} - \mathbf{\hat{b}} \Big\|^2
\label{eq:sse}
\eeqa
where 
\beqa
(\hat{\mathbf{A}},\hat{\mathbf{b}})=\arg \min_{(\mathbf{A},\mathbf{b})} \sum_{l=1}^{\Nl} \Big\| \begin{bmatrix} x^l\\y^l\end{bmatrix} - \mathbf{A} \begin{bmatrix} \hat{x}^l \\ \hat{y}^l\end{bmatrix} - \mathbf{b} \Big\|^2.
\label{eq:ab}
\eeqa

\subsection{Stopping Criterion}
\label{stop_crit}
\begin{figure}[t!]
	\centering
		\includegraphics[width=8cm,height=8cm]{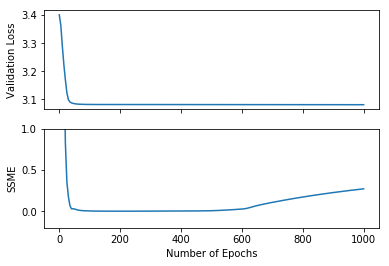}
	\caption{The validation loss (top) versus the matching error (bottom) calculated as in (\ref{eq:sse}). As it can be seen while the model has not yet overfit in terms of validation loss, it overfits in terms of matching error.}
	\label{fig:me_vs_vl}
\end{figure}

To avoid overfitting, the training must stop at some point. Early stopping based on validation data is not suitable here as the loss function (cross entropy between targets and outputs) and SSME formulated in (\ref{eq:sse}) are not directly related. In other words, overfitting in landmark positioning may happen even if the loss on validation dataset is still decreasing. This can be seen in figure \ref{fig:me_vs_vl}. As it can be seen, although the loss value on validation dataset is (slowly) decreasing (top), after a point the SSME starts to increase (bottom). In other word, although the model has not yet started to overfit in terms of loss function, it starts to overfit in terms of SSME. The point after which the model start to overfit depends on the size of the training dataset: the bigger is the training dataset size, the sooner overfit happens. Therefore we cannot choose a fixed number of epochs for training to end as it depends on training size. 

We observed that in practice, the overfit usually happens after the speed of decrease in validation loss decreases itself. And this is something which is independent of the size of training dataset. Therefore, we use the heuristic criterion which ends the training as soon as the decrease speed of loss becomes lower than a pre-specified threshold, typically a small number close to zero, e.g., $0.1$, times the biggest decrease in loss. Mathematically, denoting the validation loss at epoch $e$ with $\mathcal{L}_{e}$, the training stops when
\beq
\frac{\Delta \mathcal{L}_{e}}{\max \{\Delta \mathcal{L}_{j}\}_{j=2}^{e}}<\tau,
\label{eq:cr}
\eeq
where $\tau$ is the stopping threshold (a small number like $0.1$), and 
\beq
\Delta \mathcal{L}_{e} \triangleq \mathcal{L}_{e-1} - \mathcal{L}_e.
\label{eq:delta}
\eeq

\section{Numerical Study}
\label{numeric}

In this section, we study the performance of the proposed unsupervised landmark positioning method through numerical studies. To this end, we use synthetic data generated via simulation. We consider two different models for observed signals which models two important types of sensors used in real-life for localization.

\subsection{Pathloss Model and Received Signal Strength}
The first measurement model studied in the simulation is when the signal received by the agent is Received Signal Strength (RSS) heard from some landmarks. Landmarks in this case are radio transmitters, for example WiFi transmitters (a.k.a Access Points in the context of WiFi localization) or mobile base stations. The receiver (agent) is typically a mobile device which can measure the strength of radio signals transmitted by these WiFi Access Points (landmarks). The goal is to determine the position of landmarks only using the RSSs heard from them. This is an application which illustrates the benefits of the proposed method very well. Since WiFi access points are usually deployed for network coverage (and not positioning), therefore their exact locations is not known when we want to take advantage of them for positioning. Estimating their position using conventional methods like Weighted Centroid Localization (WCL) \cite{wcl1, wcl2}.

To model the wave propagation between a landmark, say landmark $l$, and a receiver position, say position $i$, we use the pathloss model \cite{pathloss2, pathloss3, pathloss1} as follows:
\beq
P_{i,l} = P_{T,l} - 10 n_{l} \log_{10} d_{i,l} + \eta_{i,l}
\label{eq:pathloss}
\eeq
where $P_{i,l}$ is the RSS heard from landmark $l$ at location $i$, $P_{T,l}$ is the transmit power of landmark $l$, $n_{l}$ is the pathloss exponent for landmark $l$, $d_{i,l}$ is the distance between position $i$ and landmark $l$, and $\eta_{i,l}$ is the noise.


\subsection{Inverse Linear Model}
The second model used for simulation is an inverse linear model. It is inverse linear in that the observation of the landmark is linearly proportional to its inverse distance to the measurement location. The most famous sensor obeying such a model is a camera: the size of the image of an object (landmark) in camera is (approximately) linearly proportional to its inverse distance to the camera \cite{schenk2005}.

\subsection{Experiments and Results}
{\bf Experiment 1:} The goal of the first experiment is to provide a simple visualization of the ability of the proposed method to reconstruct a map of landmarks. For the sake of visual clarity, we consider the hypothetical situation where the landmarks are equally separated on a circle. The model used for generating synthetic data is pathloss model as in (\ref{eq:pathloss}). The number of landmarks is $\Nl=30$ and the number of measurements is $\Nm=10^6$ where $80\%$ of measurements have been used for training. 

\begin{figure}[t!]
	\centering
		\includegraphics[width=7cm, height=7cm]{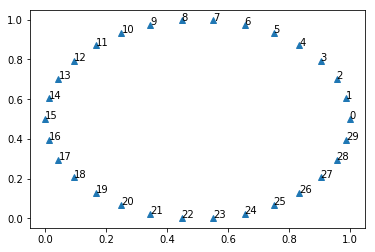}
		\includegraphics[width=8cm, height=8cm]{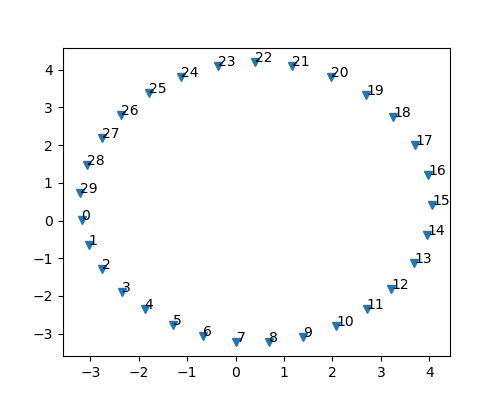}
	\caption{Reconstruction of landmark positions using Landmark2Vec: top: the true place of landmarks; bottom: their estimated positions. Each landmark has an ID between 0 and 29. As it can be seen the relative positions of landmarks have been preserved by landmark2vec. Only a scale, an orientation, and a translation differ between the two figures.}
	\label{fig:layout}
\end{figure}

The result is shown in figure \ref{fig:layout}. The training terminated when (\ref{eq:cr}) is satisfied with $\tau=0.1$. Here the training has ended only after 50 epochs with an almost perfect recovery of relative positions of landmarks. The top figure depicts the true positions of landmarks while the bottom figure depicts the estimated positions using landmark2vec. Each landmark has been specified by a label from 0 to 29. As it can be seen the relative positions of landmarks (their order on the circle) is the same in both figures. In other words, landmark2vec has retrieved the map of landmarks up to a scale, a rotation, and a translation. 

{\bf Experiment 2: } In this experiment we study a more objective measure, namely the SSME as defined in Section \ref{sec:ssme}. We compare the matching error (\ref{eq:sse}) for the two above mentioned models (pathloss and linear). The number of landmarks is $\Nl=30$ and the number of measurements is $\Nm=10^6$ where $20\%$ of measurements have been used for training. The training has ended when (\ref{eq:cr}) is satisfied with $\tau=0.025$. As it can be seen although pathloss is a more complicated model, both the SSME and the number of epochs for stopping the training are almost the same as the ones of the linear model.

\begin{figure}[t!]
	\centering
		\includegraphics[scale=0.65]{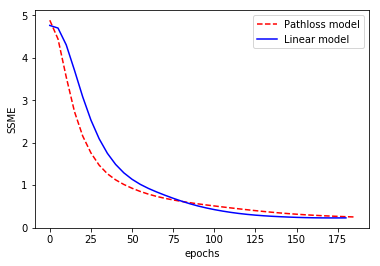}

	\caption{Reconstruction error of landmark positions using Landmark2Vec for linear and pathloss models. As it can be observed, both the SSMEs and the stopping times of the two algorithms are similar.}
	\label{fig:linear_vs_pathloss}
\end{figure}

Both experiments 1 and 2 above were implemented in Python \cite{python} version $3.6$ and Tensorflow \cite{tf} version $1.10$.

\section{Concluding Remarks}
\label{conclusion}

A neural network based method, called landmark2vec, for unsupervised positioning of landmarks was proposed, where no ground-truth information is required to estimate the position of landmarks up to a scale, rotation, and shift. The NN architecture is a shallow one comprising of just one hidden layer whose size is the same as dimensionality of space (2 for 2D positioning and 3 for 3D positioning). The training was explained and an evaluation metric was provided in order to assess the performance of the proposed method in landmark positioning. The performance was briefly illustrated and studied through numerical examples.


\end{document}